\documentclass[sn-basic,Numbered, maxcitenames=10, maxbibnames=10,]{sn-jnl}

\usepackage{graphicx}%
\usepackage{multirow}%
\usepackage{amsmath,amssymb,amsfonts}%
\usepackage{amsthm}%
\usepackage{mathrsfs}%
\usepackage[title]{appendix}%
\usepackage{xcolor}%
\usepackage{textcomp}%
\usepackage{manyfoot}%
\usepackage{booktabs}%

\usepackage{listings}%

\usepackage[detect-all=true,separate-uncertainty=true, multi-part-units=single,range-phrase={-},product-units = single]{siunitx}
\usepackage{graphicx}
\usepackage{xcolor}
\usepackage{tikz} 
\usetikzlibrary{arrows,positioning,shapes.geometric}
\usetikzlibrary{intersections}
\usepackage{import}
\usepackage[section]{placeins}
\usepackage[caption=false,font=small]{subfig}
\usepackage{wasysym}
\usepackage{booktabs}
\usepackage{dcolumn}
\usepackage{hyperref}
\usepackage{tikz}

\theoremstyle{thmstyleone}%
\theoremstyle{thmstyletwo}%

\theoremstyle{thmstylethree}%

\usepackage{hyperref}

\begin{document}

\title[Article Title]{A Mobile Robotic Approach to Autonomous 
Surface Scanning in Legal Medicine}


\author*[1]{\fnm{Sarah} \sur{Grube} \email{sarah.grube@tuhh.de } }
\author[1]{\fnm{Sarah} \sur{Latus}}
\author[1]{\fnm{Martin} \sur{Fischer}}
\author[3,2]{\fnm{Vidas} \sur{Raudonis}}
\author[4]{\fnm{Axel} \sur{Heinemann}}
\author[4]{\fnm{Benjamin} \sur{Ondruschka}}
\author[1,3]{\fnm{Alexander} \sur{Schlaefer}}

\affil[1]{\orgdiv{Institute of Medical Technology and Intelligent Systems}, \orgname{Hamburg University of Technology}, \state{Hamburg}, \country{Germany}}
\affil[2]{\orgdiv{Faculty of Electricity and Electronics}, \orgname{Kaunas University of Technology}, \state{Kaunas}, \country{Lithuania}}
\affil[3]{\orgdiv{SustAInLivWork Center of Excellence} \orgname{}, \state{Kaunas}, \country{Lithuania}}
\affil[4]{\orgdiv{Institute of Legal Medicine}, \orgname{University Medical Center Hamburg-Eppendorf}, \state{Hamburg}, \country{Germany}}


\abstract{
\textbf{Purpose}
Comprehensive legal medicine documentation includes both an internal but also an external examination of the corpse. Typically, this documentation is conducted manually during conventional autopsy. A systematic digital documentation would be desirable, especially for the external examination of wounds, which is becoming more relevant for legal medicine analysis. For this purpose, RGB surface scanning has been introduced. While a manual full surface scan using a handheld camera is time-consuming and operator dependent, floor or ceiling mounted robotic systems require substantial space and a dedicated room. Hence, we consider whether a mobile robotic system can be used for external documentation.

\textbf{Methods}
We develop a mobile robotic system that enables full-body RGB-D surface scanning.
Our work includes a detailed configuration space analysis to identify the environmental parameters that need to be considered to successfully perform a surface scan. We validate our findings through an experimental study in the lab and demonstrate the system's application in a legal medicine environment.

\textbf{Results}
Our configuration space analysis shows that a good trade-off between coverage and time is reached with three robot base positions, leading to a coverage of \SI{94.96}{\%}.
Experiments validate the effectiveness of the system in accurately capturing body surface geometry with an average surface coverage of \SI{96.90 \pm 3.16}{\%} and \SI{92.45 \pm 1.43}{\%} for a body phantom and actual corpses, respectively. 

\textbf{Conclusion}
This work demonstrates the potential of a mobile robotic system to automate RGB-D surface scanning in legal medicine, complementing the use of post-mortem CT scans for inner documentation. 
Our results indicate that the proposed system can contribute to more efficient and autonomous legal medicine documentation, reducing the need for manual intervention.
}



\maketitle
\section{Introduction}\label{sec1}
In legal medicine, documentation of corpses is a common and increasingly important task, serving multiple purposes such as documenting injuries for medico-legal investigations and providing evidence for legal proceedings. 
To achieve comprehensive documentation, it is essential to analyze not only the internal structures of the body, but also to accurately document the external surface of the body. By integrating surface and internal structure data, the aim is to create comprehensive legal medicine records \cite{Thali.2005, UrsulaBuck.2013,  Campana.2016, Ebert.2010}.

While post-mortem computed tomography (PMCT) and post-mortem magnetic resonance imaging (PMMRI) provide detailed insights into internal anatomy and pathology, these modalities are limited in capturing detailed surface information, such as skin color and the finer details of surface injuries. Therefore, body surface analysis is decisive in accurately documenting the external shape and color of abnormalities such as wounds. 

Various system approaches have been used for documentation \cite{Buck.2018}. 
Conventionally, 2D imaging using digital cameras has been used. However, advances in technology have led to more sophisticated methods, particularly 3D surface scanning techniques. Photogrammetry, for example, uses multiple images taken from different angles with a handheld camera, camera rig or a special rod to create a 3D model \cite{PetraUrbanova.2015, Kottner.2017, Kottner.2021, TillSieberth.2024}. 
Alternatively, 3D surface scanning with hand-held time-of-flight cameras is used. These approaches offer mobility and flexibility, allowing the scanning system to be used in different rooms and environments, hence not limiting the system to a single fixed configuration.
Despite these advances, there are trade-offs within these approaches, including cost efficiency, mobility, repeatability and automation. Cost-effective approaches typically require manual and well-trained operation with a lack of automation. 
Automated approaches tend to be expensive and require a fixed setup to ensure consistent positioning of the scanned body relative to the system \cite{Ebert.2010, Kottner.2017}.
A cost-effective approach is presented in Sieberth et al.~\cite{TillSieberth.2024}, using a spoke rod with multiple cameras. However, manual scanning is required. 

There is growing interest in automating these procedures using robotic systems~\cite{Nawrat.2023} to make the documentation process more time efficient and to minimize the risk of infection. The swiss Virtopsy movement~\cite{Thali.2005} focuses on providing high-resolution multi-modal 3D color documentation of corpses. 
As part of the Virtopsy project, the Virtobot was developed~\cite{Ebert.2010} for automated surface scanning using predefined robot positions. This system already shows promising results at the Institute of Forensic Medicine in Zurich.
However, the high costs, a fixed large installation with specific ceiling requirements to mount the robot, the required thorough preparation of the body, and the application of multiple markers to the body for accurate data stitching make this setup less flexible. Together with the necessary manual path planning of the robot, these requirements limit the autonomy of the system.

In summary, the need for automated, time-efficient and multi-modal solutions for legal medicine documentation is growing and increasingly important. Such solutions should be low-cost and mobile with respect to different space constraints and autonomous to reduce the workload and minimize the risk of infection.

In this article, we present a mobile and autonomous surface scanning robot for comprehensive external examination and digital documentation of corpses, including for complementing PMCT scanning or pre-autopsy documentation. In our context, autonomous refers to a robot that can recognize the pose of a body, independently decide which positions are best for RGB-D surface acquisition, and independently perform the scanning process.
A robot base with differential drive enables flexible positioning of a lightweight 6-joint robot arm with an RGB-D camera attached for surface scanning.
The mobile base allows us to increase mobility and adaptability by extending the operating range of a robotic arm, providing a more autonomous alternative to traditional approaches. 
The RGB-D information is used both for autonomous control and navigation of the robotic setup in the current environment and for surface scanning.

Our experimental evaluation aims to investigate the performance of our system for comprehensive surface scanning. 
First, we investigate which environmental settings have to be considered in order to successfully perform a surface scan and how many different robot base positions are needed. We use a configuration space analysis, to perform a detailed parametric analysis, investigating factors such as different workspace constraints, body geometries, and camera positioning, as well as the trade-off between scanning time and surface resolution. 
Second, we validate our results with real-world experiments in a laboratory setting and demonstrate the application of our system in legal medicine.

\section{Material and Methods}\label{sec11}

\begin{figure}[tb]
	\centering
    \subfloat[General Scanning Setup]{
        {\includegraphics[width=0.37\linewidth]{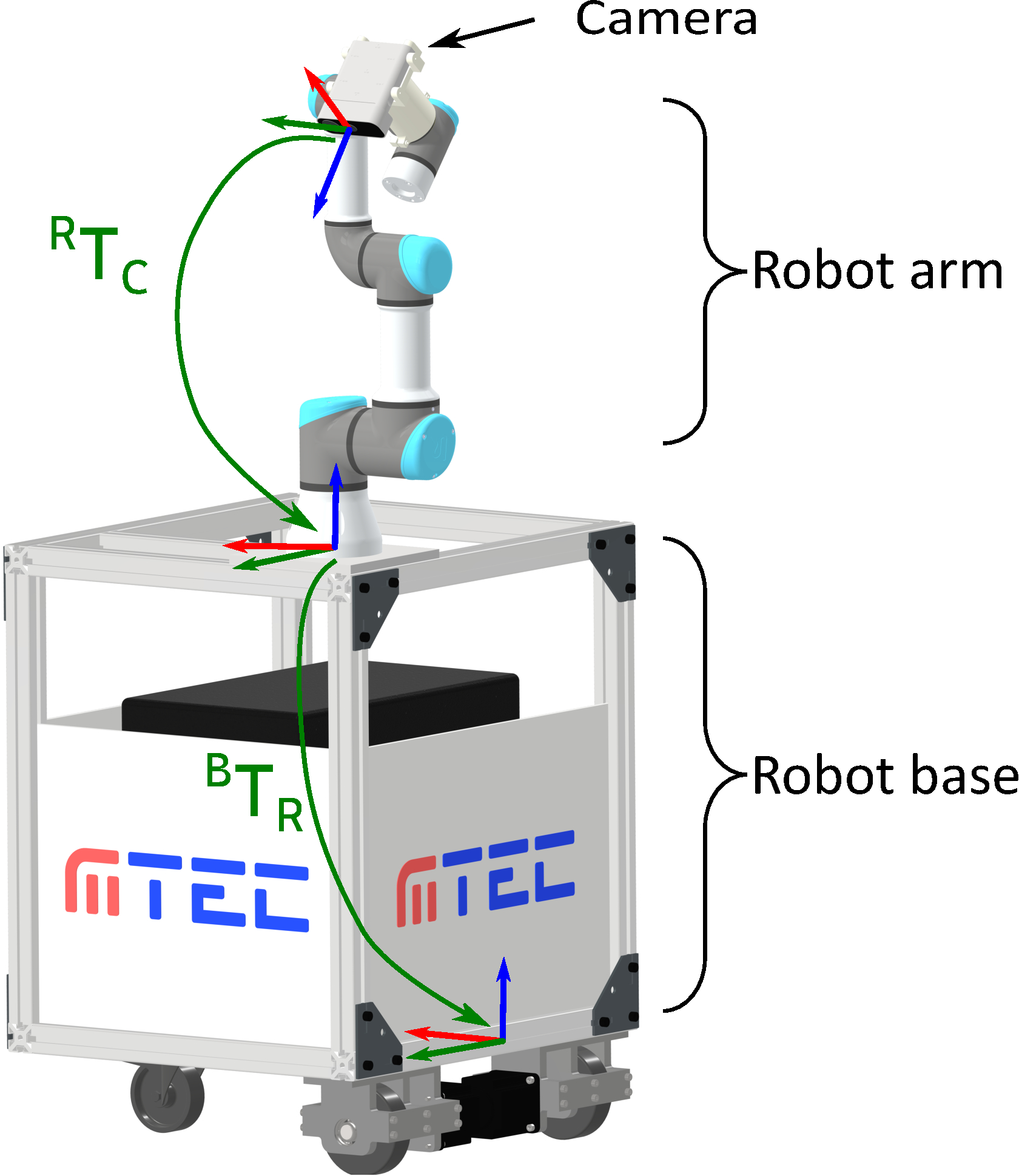}}
        \label{fig:generalLabSetup}
    }
    \hfill
    \subfloat[Scanning Workflow]{
        \includegraphics[width=0.58
        \linewidth]{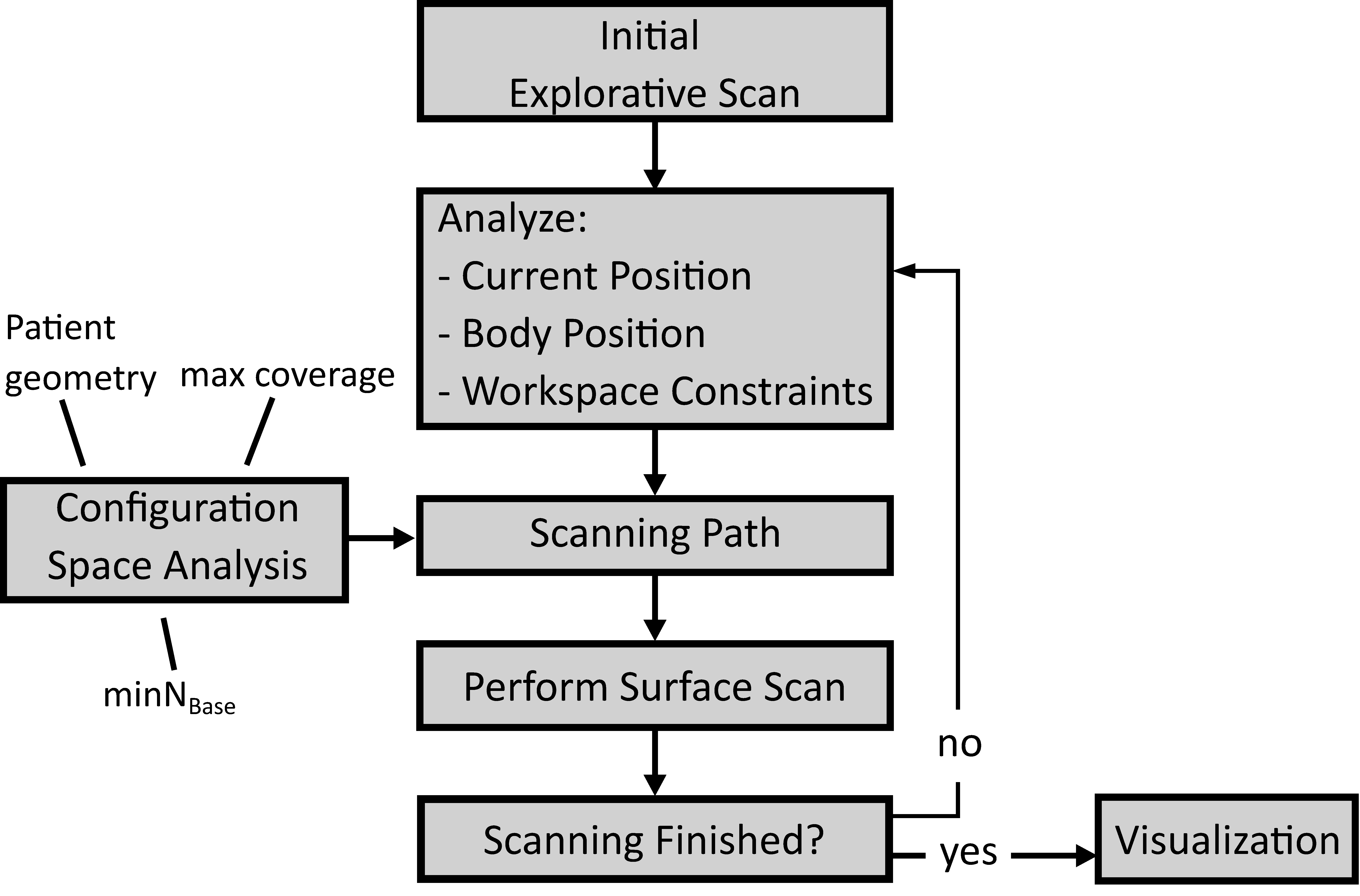}
        \label{fig:GeneralWorkflow_hochkant}
    }
	\caption{\textbf{General System Setup:} Schematic drawing of the mobile scanning setup (left) and the corresponding scanning workflow (right)}	
	\label{fig:simulationProcedure}
\end{figure}

\subsection{General Setup and Workflow}
\label{sec:SetupAndWorkflow}
Our mobile robotic system depicted in Figure~\ref{fig:generalLabSetup} is designed to enhance autonomous surface scanning. 
The system consists of a mobile base that allows flexible positioning of a robotic arm around the body. We refer to the mobile base as robot base.
The robotic arm is equipped with an RGB-D camera for surface scanning and navigation. 
The proposed workflow of an autonomous 3D surface scanning procedure using our system is shown in Figure~\ref{fig:GeneralWorkflow_hochkant}. 

A configuration space analysis is conducted once, based on the given robotic system parameters, to define scanning paths that ensure maximum coverage of a body geometry while minimizing the number of robot base positions.
Our scanning workflow starts with an initial explorative scan to assess relevant environmental factors such as the initial orientation of the patient and robot, spatial constraints such as obstacles, couch height and body position. 
Subsequently, the initial explorative scan is combined with the conducted configuration space analysis to select an optimal scanning path considering the identified environmental constraints.
Path planning for the mobile robot base is performed using an A* algorithm~\cite{Hart.1968}. Surface scanning is performed by iteratively moving the robot base and robot arm following the predefined optimal scanning path. 
Once the body has been scanned, post-processing steps are performed and the surface scan is displayed to the clinician.

\subsection{System Parameters and Calibration}
Our autonomous scanning system consists of a lightweight small UR3~(Universal Robots, DK) robotic arm mounted on a differential drive robot base. An RGB-D camera~(Azure Kinect,  Microsoft, USA) is attached to the fifth joint of the UR3. 
We use the Robot Operating System~(ROS) to control the whole system and limit the velocity of the robot base to \SI{0.08}{m/s}.

Initially, hand-eye calibration using Parks method~\cite{Park.1994} is performed to obtain the transformation $^RT_C$. Using an 8x5 tile checkerboard, we acquire 100 images for random robot arm configurations. We obtain reprojection errors of \SI{0.039} and \SI{0.271}{px} for RGB and IR imaging, respectively. 
For hand-eye calibration we obtain a translation error of \SI{6.21}{mm} (x= \SI{2.26}{mm}, y= \SI{4.72}{mm}, z= \SI{2.34}{mm}) and a rotational error of \SI{0.423}{°}. 
We define the transformation $^BT_R$ between the UR3 base and the robot base based on the kinematic design.

\subsection{Configuration Space Analysis}
We aim to systematically identify the necessary robot configurations to enable a comprehensive scanning of a specific body geometry. A configuration space analysis is performed that optimizes with respect to both the maximum coverage of the body surface and the minimum number of robot base positions used. The configuration space of the scanning robot is defined by the position of the robot base and the configuration of its arm. 
We assume the best visibility when the camera is orthogonal to the body surface. Consequently, for each robot base position, we determine which body surface points can potentially be seen and which configurations of the robot arm are required.
Finally, the configuration space analysis provides a dictionary of all configurations and the corresponding visible points.

\subsection{Selecting the Scanning Path} 
In a subsequent step, the dictionary from the configuration space analysis is used to identify the most promising robot base and robot arm configurations for scanning the body surface.
This process uses a greedy algorithm to iteratively add the robot configuration with the largest number of currently unseen surface points until no new points can be captured.
Note that changing the camera field of view by moving the robot arm is faster than changing the robot base position. Therefore, we first move the robot arm to reach unseen surface points.

\subsection{Reconstruction of Surface Scan}
A full body surface surface scan is reconstructed by stitching the acquired point clouds based on the respective robot configurations. We first merge the captured point clouds for the individual robot base positions and then arrange these in space.
Finally, a point-to-point Iterative Closest Point~(ICP) algorithm is used for fine alignment with subsequent outlier removal. All steps are implemented using the Python open3d library. To reduce computation time, point clouds are down-sampled to a resolution of \SI{10}{mm}.

\subsection{Experimental Evaluation}
The experimental evaluation is divided into a computational analysis of the configuration space and corresponding scanning paths and two experimental studies applying the robotic setup in both a laboratory and a clinical setting.

\subsubsection{Configuration Space and Path Evaluation}
A detailed configuration space analysis is performed to investigate the influence of different environment settings and to determine appropriate parameter settings for a comprehensive surface scan. 

To decide which point cloud resolution to use for the configuration space analysis, we investigate the trade-off between point cloud resolution, coverage and run time, using an average male surface model~\cite{Schlehlein2022}.
In particular, we consider resolutions ranging from \SI{0.5}{m} to \SI{0.025}{m}. We use the kneedle algorithm~\cite{Satopaa2011FindingA} to find an optimal trade-off.

The environmental parameters studied are body geometry, couch height, and workspace constraints.
First, we study the influence of different body geometries on the scanning coverage, namely half-cylinder, average male mesh, and surfaces extracted from CT scans. In particular, we use 10 different corpse surfaces extracted from CT scans acquired at the Institute of Legal Medicine at the University Medical Center Hamburg-Eppendorf~(Legal Medicine, UKE). Please note, that we use a half-cylinder approach for path planning since the body geometry is usually not known before scanning. For all body models, we neglect workspace constraints and use a surface point cloud resolution of \SI{0.1}{m} with a couch height of \SI{0.67}{m}. 

In subsequent evaluations, we only use the CT scans as body models. 
Second, different couch heights are investigated, considering values of \SI{0.67}{m} and \SI{0.9}{m}, as both heights are within the range of CT table heights and autopsy table heights.

\begin{figure}[tb]
	\centering
    \subfloat[Full Workspace]{
        \includegraphics[width=0.3\linewidth]{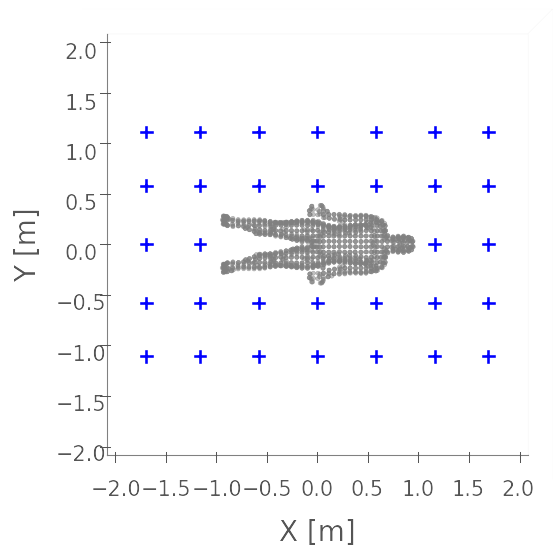}
        \label{fig:wholeWorkspace}
    }
    \subfloat[Narrow Workspace]{
        \includegraphics[width=0.3\linewidth]{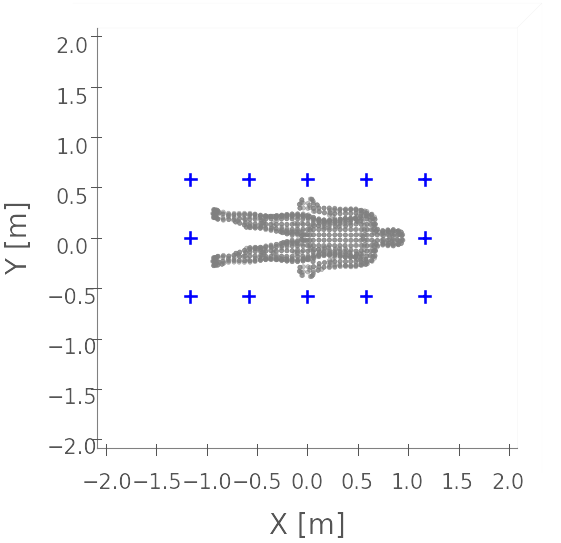}
        \label{fig:narrowWorkspace}
    }
    \subfloat[Access from one side only]{
        \includegraphics[width=0.3\linewidth]{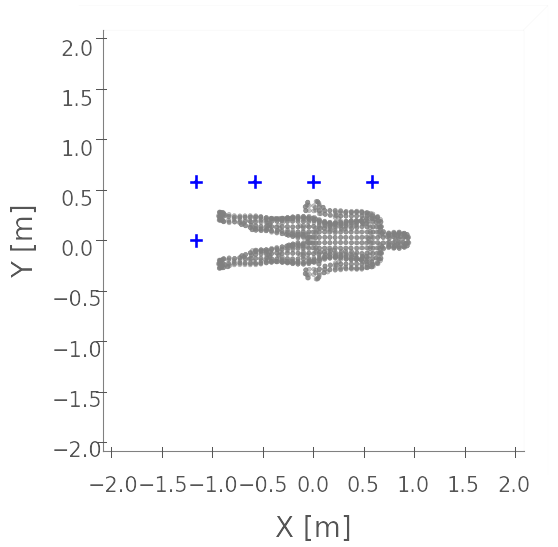}
        \label{fig:FMWorkspace}
    }
	\caption{\textbf{Workspace Constraints:} different workspace constraints investigated in the experiments. Full workspace (left), narrow room (middle) and access from only one side (right)}	
	\label{fig:simWorkspace}
\end{figure}
Third, we evaluate the effect of workspace constraints visualized in Fig.~\ref{fig:simWorkspace}. Confined environments are considered, as well as environments in which the body is only accessible from one side. Both environments correspond to typical constraints in the clinics.

For comparability, we always calculate the coverage for the corresponding high-resolution~(\SI{0.01}{m}) point cloud and do not take into account points from the back of the body, as they are hidden and therefore not accessible. 
A maximum of five different robot arm configurations per robot base position are considered to limit the scan time.
We used the Python RoboticsToolbox and the MoveIt package for the computational analysis. The code is executed on an Intel Core i7-13700K CPU with 64GB of RAM.

\subsubsection{Real World Evaluation} 

\begin{figure}[tb]
	\centering
    \subfloat[Lab Environment]{
        \includegraphics[width=0.4\linewidth]{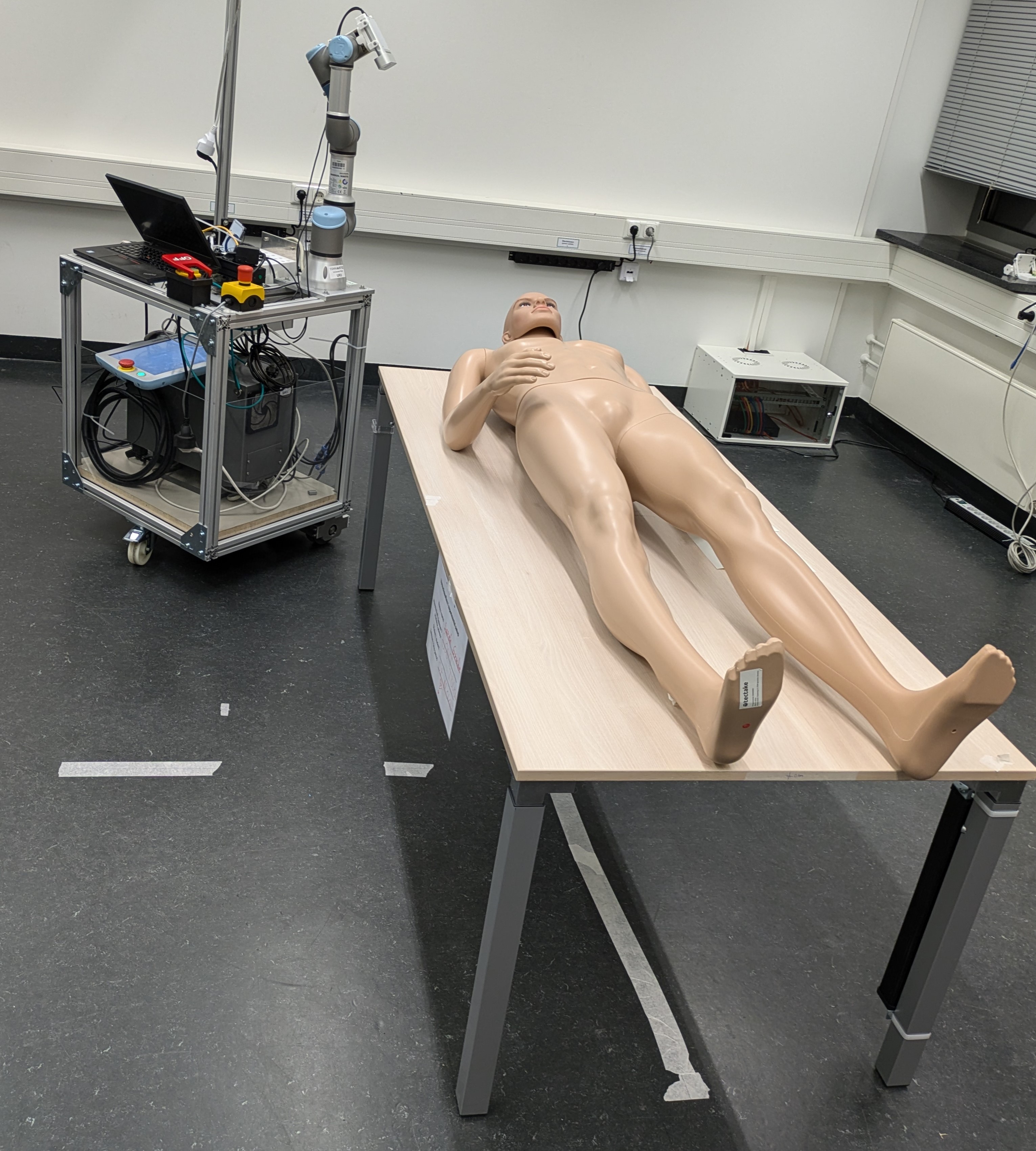}
        \label{fig:labSetup_mobileBase_Plus_Mannequin}
    }
     \subfloat[Robot Base Positions]{
        \includegraphics[width=0.5\linewidth]{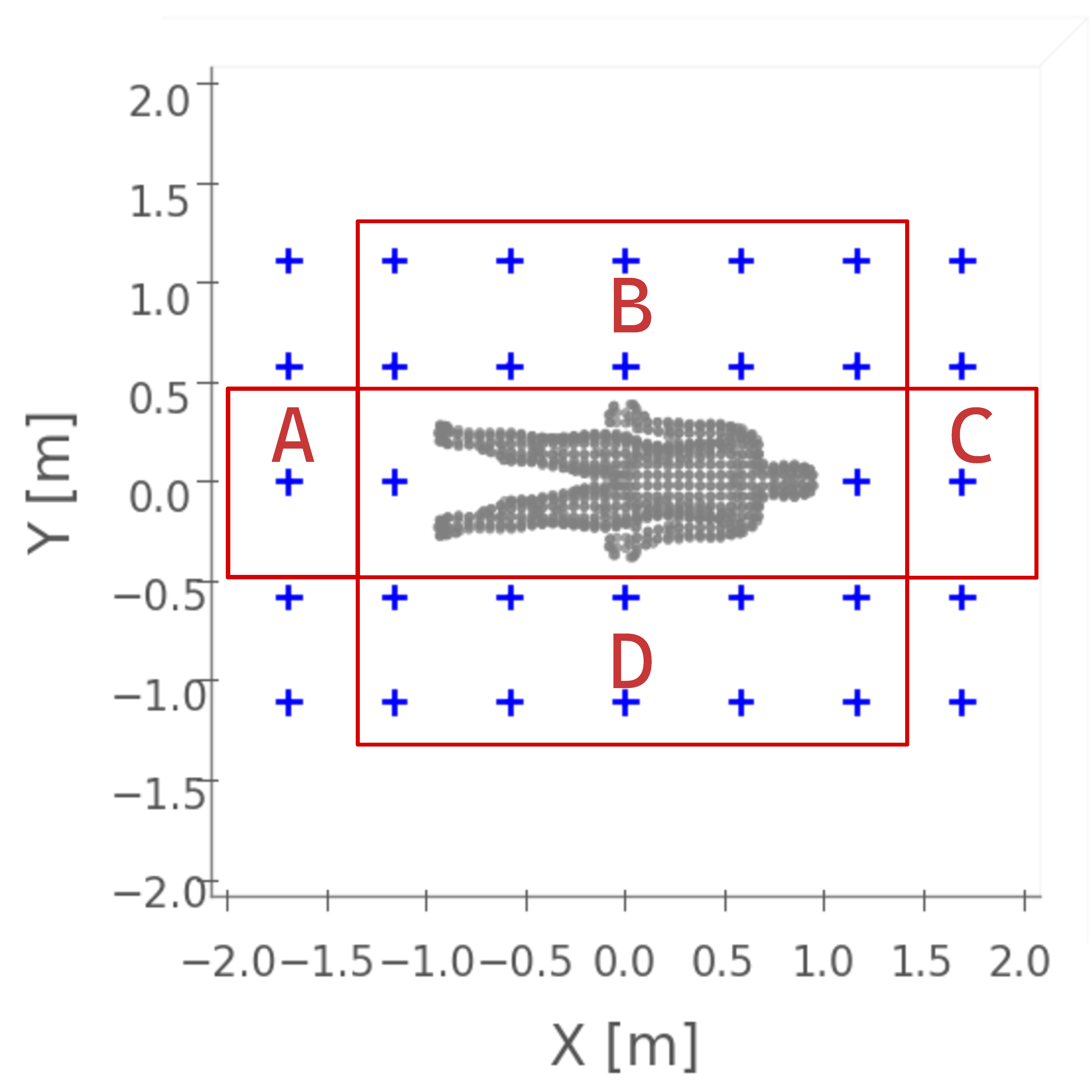}
        \label{fig:labStartpositions}
    }
	\caption{\textbf{Lab:} Lab environment (left) with autonomous scanning system and body phantom used for surface scanning. The investigated areas of starting positions for surface scanning are shown in (b)
}

	\label{fig:simulationProcedure}
\end{figure}

We evaluate the performance of our surface scanning approach in two real-world environments. More specifically, we use our robot system in a laboratory setting without workspace constraints and in a forensic CT room with constraints similar to the evaluations in Fig.~\ref{fig:simWorkspace}. In all scenarios, the robot is positioned randomly in the room. 
A maximum of three additional robot base positions per scan are used with a maximum of five robot arm configurations each. 

In the laboratory setting~(Fig.~\ref{fig:labSetup_mobileBase_Plus_Mannequin}), we use a male body phantom as scanning object. To serve as a reference, the surface geometry of the body phantom is acquired using a hand scanner~(Artec Eva, Artec 3D, LU). Please note that we perform several manual post-processing steps on the captured point clouds, such as manual fragment removal, fusion of multiple point clouds, and smoothed fusion. These steps are performed using the Artec Studio 17 Professional software.

To examine the repeatability of our system as a function of the initial robot base position in our laboratory setting, we divide the working region into four areas as defined in Fig.~\ref{fig:labStartpositions}. For each area, the robot base is randomly positioned and a surface scan is performed. This process is repeated ten times for each area, resulting in 40 surface scans with different starting positions. We evaluate our acquired surface scan using the coverage achieved compared to the reference scan from the hand scanner. To determine the coverage achieved, we compare the acquired surface scan with the registered surface from the handheld scanner. Here, we assume that a point is covered if a corresponding point in the handheld scan is available and the registration error is less than twice the voxel size.\\

Additionally, to the experiments performed in the laboratory setting, we evaluated our system in Legal Medicine. We scanned three male corpses and validated the acquired surface scan with the surface extracted from a corresponding CT scan~(Incisive CT, Philips Healthcare, Netherlands). We used Slicer for surface extraction from the CT volume based on thresholding. 
Note that parts of the body were covered for anonymity. For evaluation, we calculate the coverage reached compared to the CT surface.

\section{Results}\label{sec2}
\subsection{Configuration Space and Path Evaluation}
The results of the tradeoff analysis between point cloud resolution and runtime of the configuration space analysis are detailed in Table~\ref{tab:pcdRes_CSAtime}. An optimal solution is reached with a resolution of \SI{0.1}{m}, yielding \SI{96.43}{\%} surface coverage and requiring \SI{3.44}{min} for analysis of one robot base position.\\

\begin{table}[tb]
\begin{minipage}{1\linewidth}
\caption{\textbf{Configuration Space Analysis:} Impact of the point cloud resolution used in the Configuration Space Analysis on the coverage of the surface scan. Additionally, the corresponding mean time~$t_\text{BP}$ for the configuration space analysis for a single robot base position in minutes is given}

\label{tab:pcdRes_CSAtime}%
\begin{tabular}{lll}
    \toprule
    \textbf{Resolution [$m$]} & \textbf{Surface Coverage [\%]} & \textbf{$t_\text{BP}$ [min]} \\
    \midrule
    0.025 &98.14 &58.47 \\
    \textbf{0.1} &\textbf{96.43} &\textbf{3.44}  \\
    0.25 &86.39 &0.69  \\
    0.5 &62.23 &0.25  \\
    \botrule
    \end{tabular}
\end{minipage}
\end{table}

The quantitative results of our configuration space analysis are shown in Table~\ref{tab:CSA_Parameteranalysis}. Assuming a half-cylinder body geometry, full coverage is achieved using four robot base positions. Comparing the achieved coverage between three and four robot base positions, only a slight increase of \SI{1}{\%} is observed.

Using a half-cylinder body model for path planning, a coverage of \SI{96.10}{\%} was achieved for the average male mesh, while the different CT surfaces resulted in an average coverage of \SI{93.80 \pm 3.15}{\%}. 

Considering the detailed body geometry for path planning, similar results were obtained with a coverage of \SI{96.43}{\%} and \SI{94.96 \pm 3.02}{\%} for average male mesh and CT surfaces, respectively.

Variations in couch height showed that increasing the height to \SI{0.90}{m} resulted in a \SI{3.43}{\%} reduction in coverage.
Workspace constraints limiting access to a single couch side reduced the potential surface coverage by \SI{10}{\%}, leaving coverage of \SI{84.92}{\%}. A narrow room results in similar coverage to no workspace constraints.

\begin{table}[tb]
\begin{minipage}{1\linewidth}
\caption{\textbf{Configuration Space Analysis:} Influence of different parameters on the achieved surface coverage $[\%]$. The surface coverage for 1-4 different robot base positions is given}
\label{tab:CSA_Parameteranalysis}%
\begin{tabular}{lcccc}
    \toprule
    \textbf{} & \multicolumn{4}{c}{ \textbf{Number of base positions} }\\
    \toprule
    \textbf{Parameter} & \textbf{1} & \textbf{2} & \textbf{3} & \textbf{4} \\
    \midrule
    Half-cylinder  &70.91 &95.81 &99.60 &100.00\\
    Average male mesh&75.97 &91.01 &94.29 &96.10 \\
    CT surface  &$69.48 \pm 6.49$ &$89.60 \pm 4.28$ &$92.88 \pm 3.50$ &$93.80 \pm 3.15$\\
    \midrule
    Couch height (\SI{0.67}{m})  & $75.05 \pm 3.92$ &$91.02 \pm 2.71$ &$93.78 \pm 2.61$ &$94.96 \pm 3.02$\\
    Couch height (\SI{0.9}{m}) &$58.25 \pm 6.69$ &$83.97 \pm 5.13$ &$89.92 \pm 3.77$ &$91.53 \pm 3.26$\\
    \midrule
    Narrow Room &$72.93 \pm 3.97$ &$89.42 \pm 3.16$ &$93.53 \pm 3.35$ &$93.70 \pm 3.32$\\
    One side only& $73.53 \pm 4.01$ &$81.22 \pm 4.84$ &$84.55 \pm 4.46$ &$84.92 \pm 4.23$\\
    \botrule
    \end{tabular}
\end{minipage}
\end{table}

\subsection{Real World Evaluation} 

\begin{figure}[tb]
	\centering
 \hfill
    \subfloat[Hand scanner RAW]{
        \includegraphics[width=0.15\linewidth]{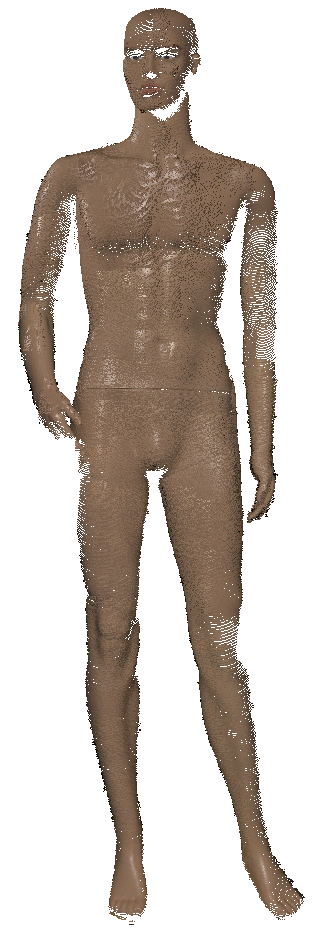}
        \label{fig:handscannerRAW}
    }
    \hfill
    \subfloat[Hand scanner smoothed]{
        \includegraphics[width=0.145\linewidth]{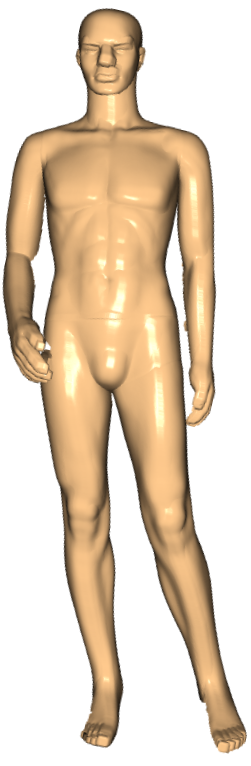}
        \label{fig:handscannerPP}
    }
    \hfill
    \subfloat[Single Base Position]{
        \includegraphics[width=0.162\linewidth]{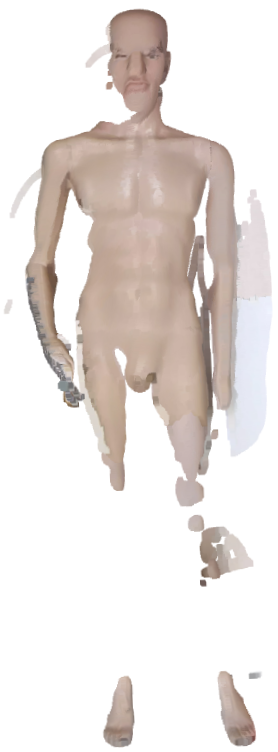}
        \label{fig:CT_ID3_GT}
    } 
    \hfill
    \subfloat[Stitched Surface Scan]{
        \includegraphics[width=0.157\linewidth]{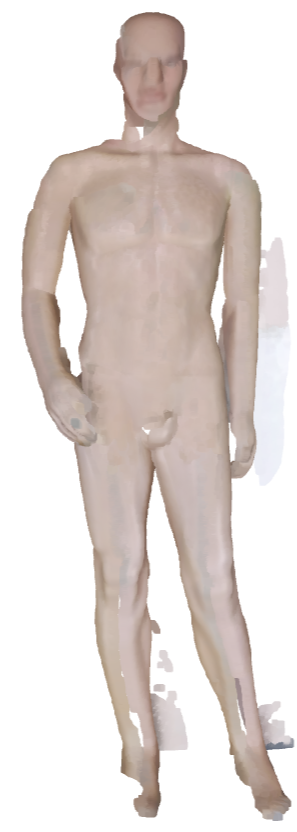}
        \label{fig:CT_ID3_GT}
    } 
    \hfill
     \subfloat[Distance to Hand-scanned Surface Scan]{
    \includegraphics[width=0.243\linewidth]{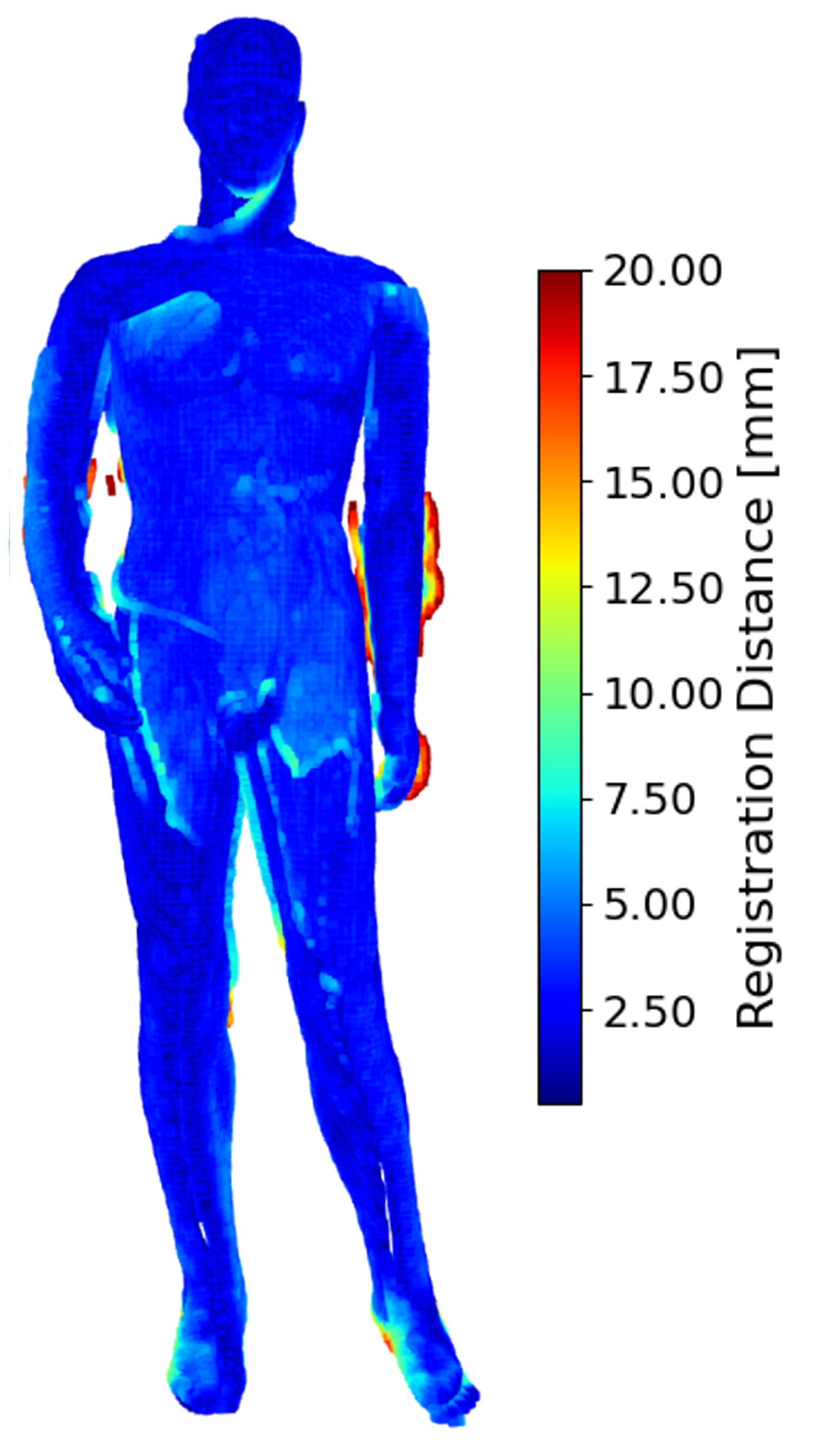}
        \label{fig:labStitched_error}
        }
	\caption{\textbf{Lab:} Example of an acquired surface scan after stitching. The reference surface acquired with the hand scanner~(a) is shown as well as a surface scan using only one robot base position (b) and using all robot base positions (c).
    In (d) a color comparison regarding the absolute surface distance to the hand-scanned reference surface is shown}	
	\label{fig:Lab_acquiredPCD_vsHS}
\end{figure}

The results of our experiments performed in the lab are shown in Table~\ref{tab:lab_NumPosVsCoverage}. We reach a mean coverage of \SI{93.78 \pm 2.61}{\%} with a mean difference of \SI{6.01}{mm} compared to the hand-scanned reference surface. Similar coverage is achieved when evaluating the different starting positions (A-D).
An exemplary surface scan is visualized in Figure~\ref{fig:Lab_acquiredPCD_vsHS}. The distance differences to the registered hand-scanned surface are indicated with color coding in Figure~\ref{fig:labStitched_error}.

\begin{figure}[tb]
	\centering
    \subfloat[]{
        \includegraphics[height=4.5cm]{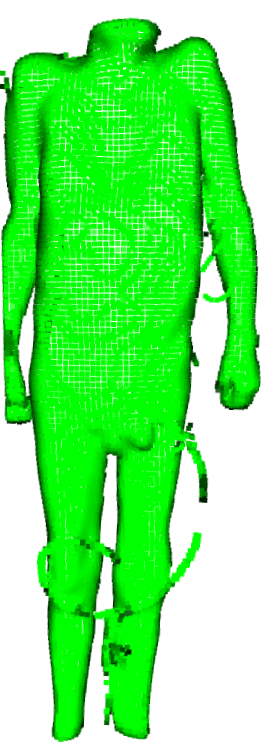}
        \label{fig:pat2_CT_GT}
    }
    \hfill
    \subfloat[]{
        \includegraphics[height=4.5cm]{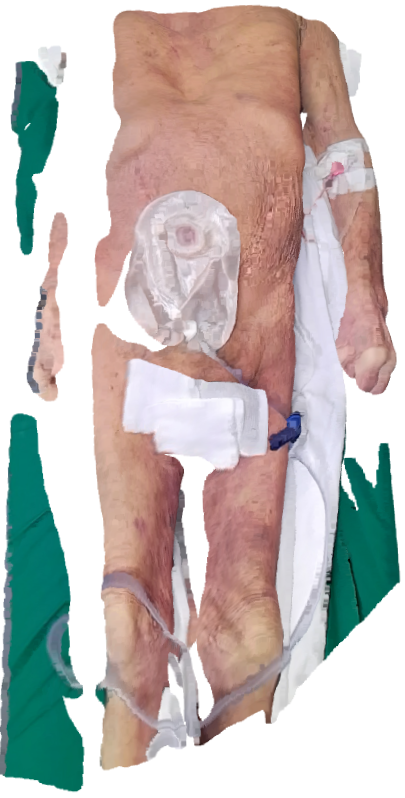}
        \label{fig:pat2_kinectScan_single}
    }
    \hfill
    \subfloat[]{
        \includegraphics[height=4.5cm]{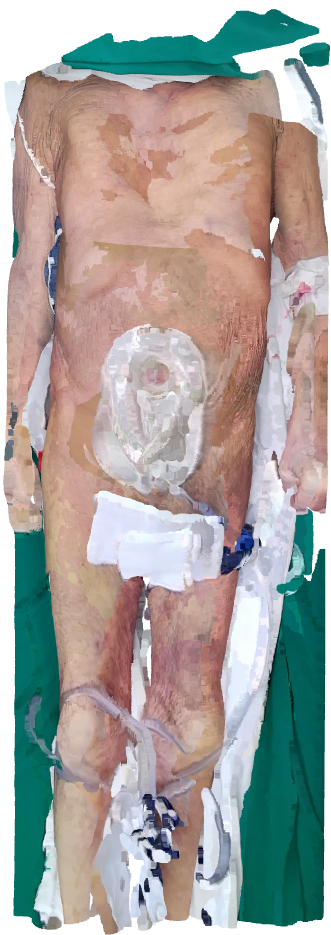}
        \label{fig:pat2_kinectScan}
    } 
    \hfill
    \subfloat[]{
        \includegraphics[height=4.5cm]{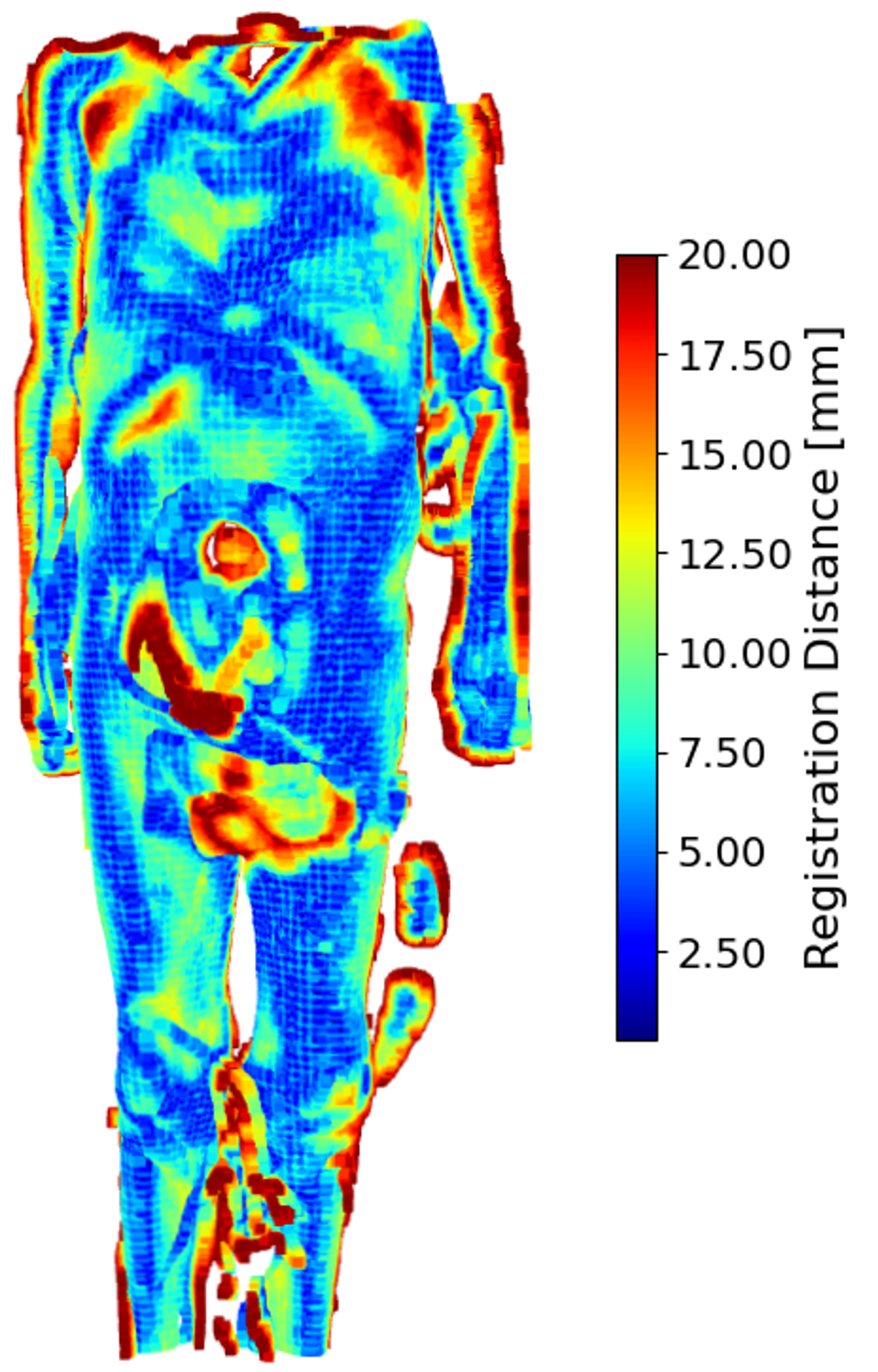}
        \label{fig:pat2_distCTReg}
        }
    \newline
    \subfloat[]{
        \includegraphics[height=4.5cm]{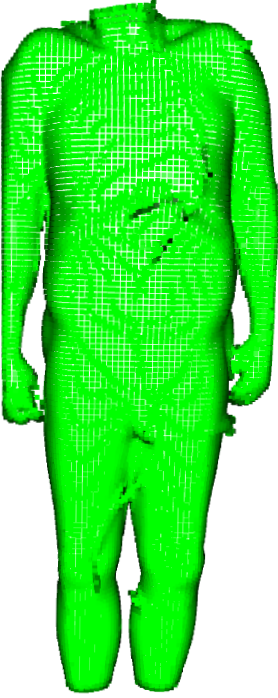}
        \label{fig:pat3_GT}
    } 
    \hfill
    \subfloat[]{
        \includegraphics[height=4.5cm]{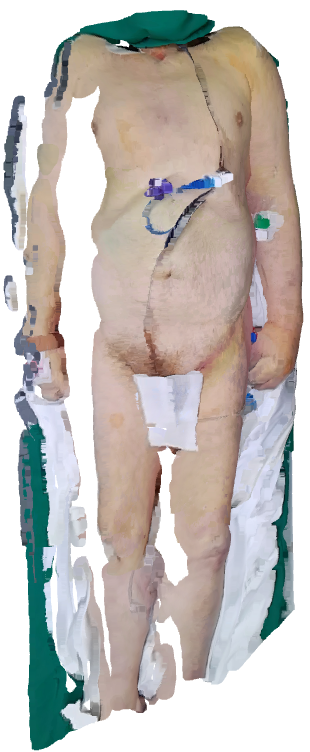}
        \label{fig:pat3_kinectScan_single}
    } 
    \hfill
    \subfloat[]{
        \includegraphics[height=4.5cm]{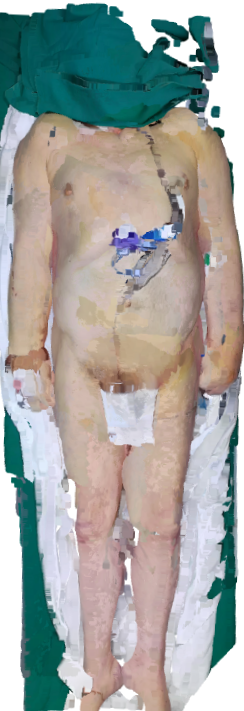}
        \label{fig:high_res_stitchedSurface_ID3}
    }
    \hfill
    \subfloat[]{
        \includegraphics[height=4.5cm]{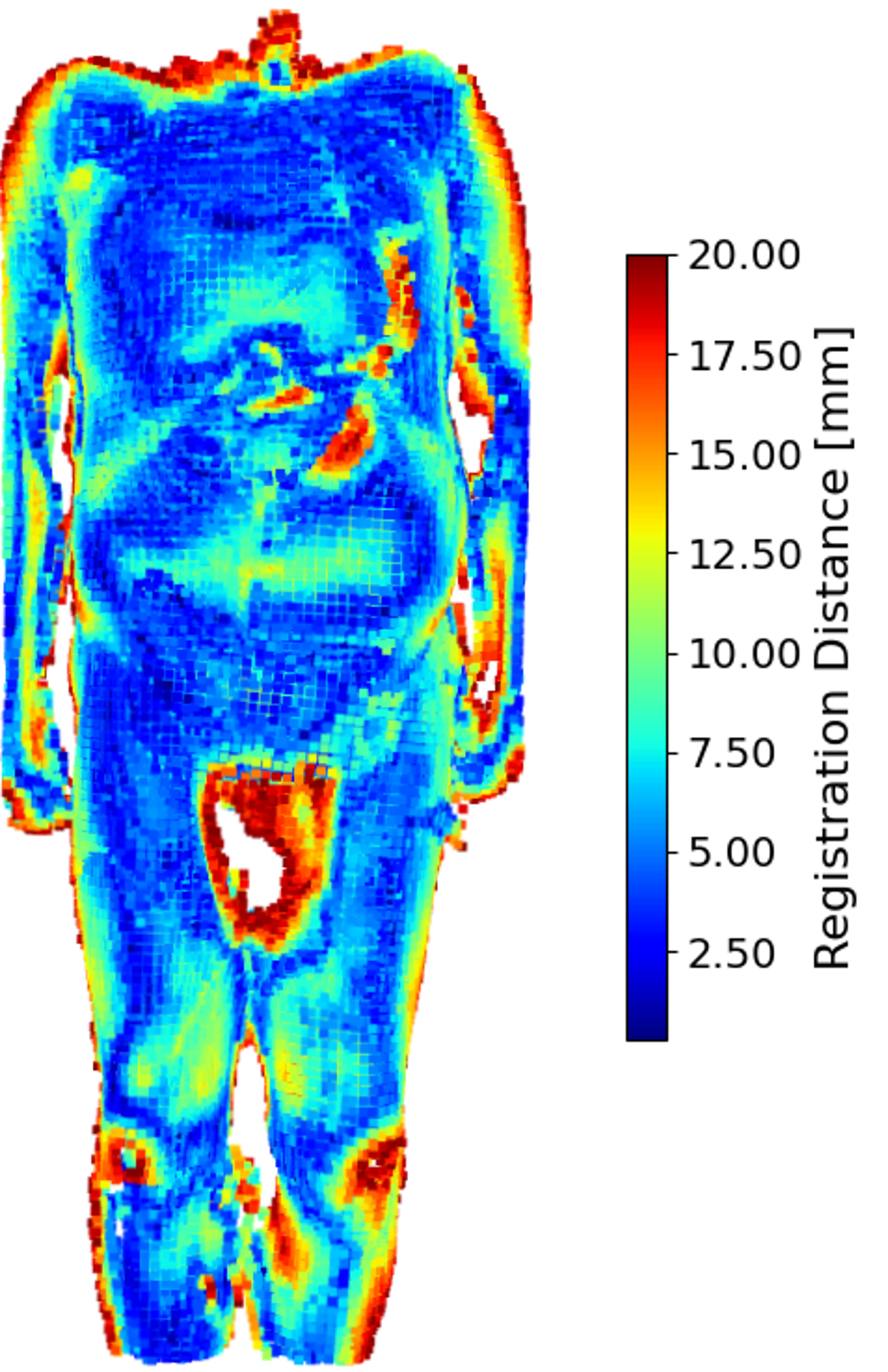}
        \label{fig:FMStitched_P3_distancetoCT_WithColorbar}
        }
    \newline
     \subfloat[]{
        \includegraphics[height=4.5cm]{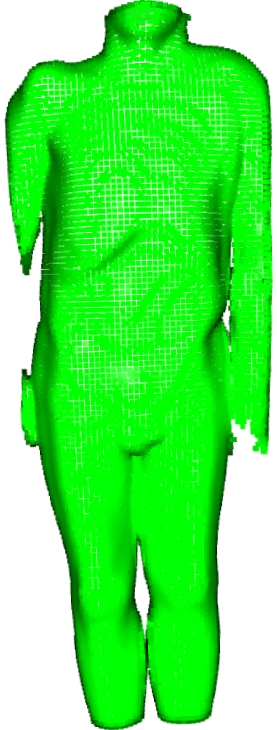}
        \label{fig:pat1_GT}
    } 
    \hfill
    \subfloat[]{
        \includegraphics[height=4.5cm]{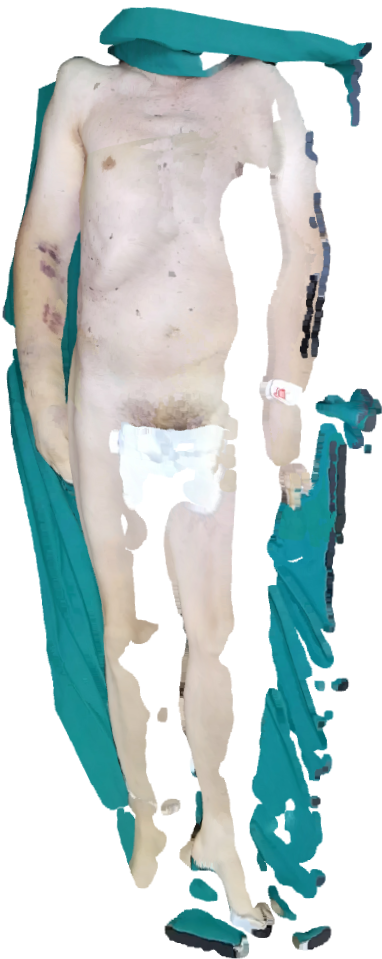}
        \label{fig:pat1kinectScan_single}
    } 
    \hfill
    \subfloat[]{
        \includegraphics[height=4.5cm]{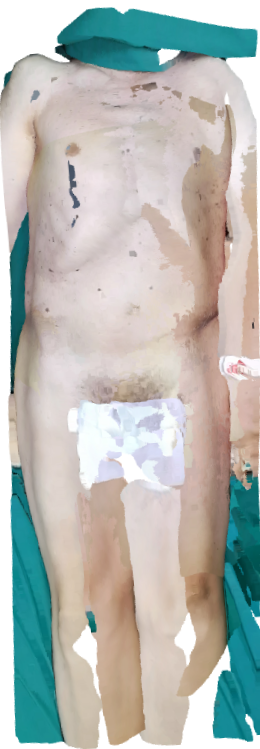}
        \label{fig:pat1_kinectScan}
    }
    \hfill
    \subfloat[]{
        \includegraphics[height=4.5cm]{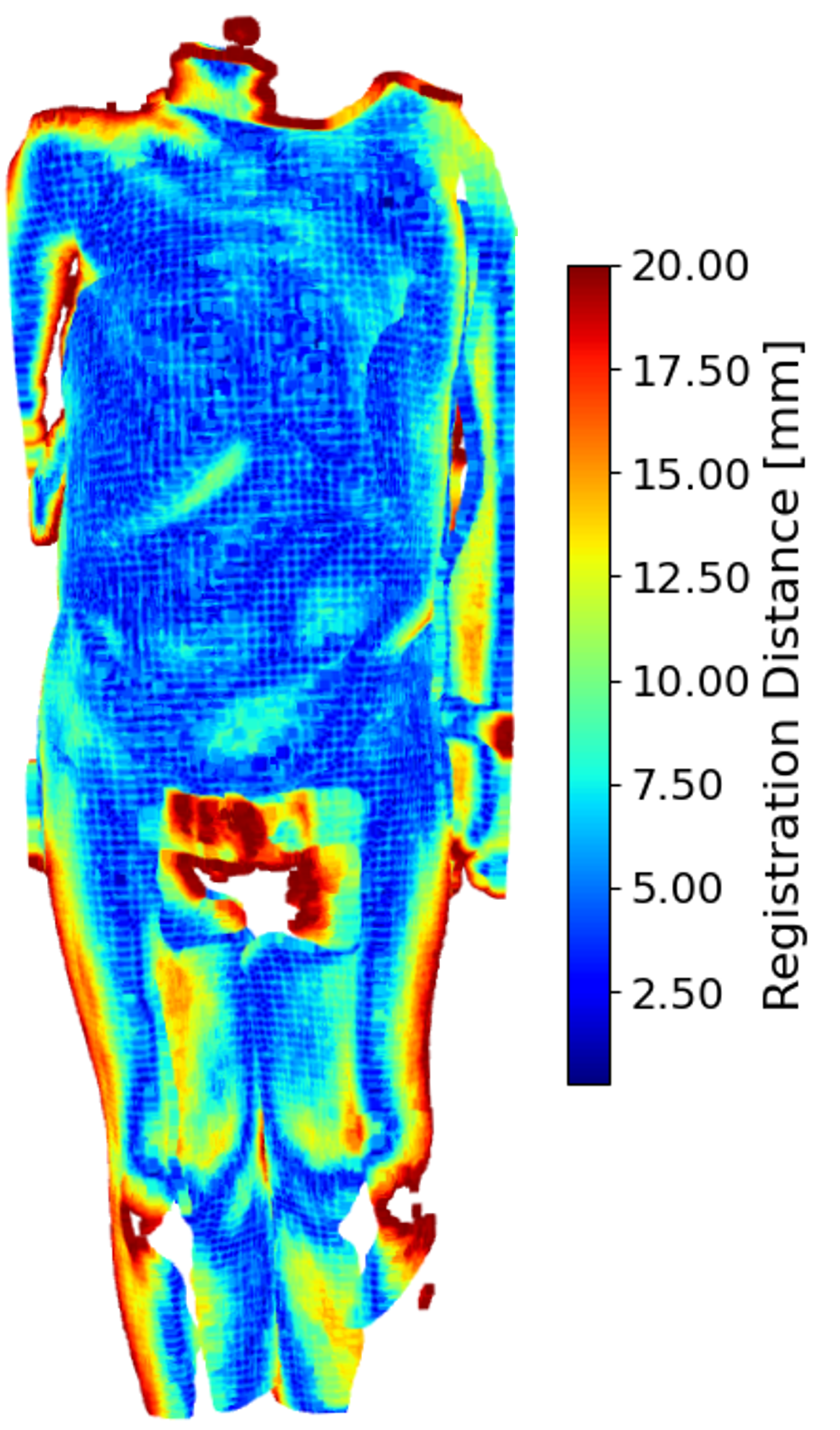}
        \label{fig:pat1_distCTReg}
        }     
	\caption{\textbf{Legal Medicine:} acquired surface scans for three different corpses. (a,e,i) show the surface extracted from the CT scan, (b,f,j) a surface scan from a single robot base position and (c,g,k) the surface scan using several robot base positions. In (d,h,l) a color comparison regarding the absolute surface distance to the registered CT-surface is shown}	
	\label{fig:FM_acquiredPCD_vsCT}
\end{figure}

The surface scans of three corpses acquired at Legal Medicine are shown in Figure~\ref{fig:FM_acquiredPCD_vsCT}. Each row corresponds to a corpse and shows the extracted surface from the CT scan, an acquired surface scan from a single robot base position, the entire surface scan, and a color comparison regarding the absolute surface distance to the registered CT surface. The three surface scans reach a coverage of \SI{90.80}{\%}, \SI{93.30}{\%}, \SI{93.26}{\%} and a mean registration distance of \SI{7.44}{mm}, \SI{7.34}{mm}, \SI{7.51}{mm}.

\begin{table}[tb]
\begin{minipage}{1\linewidth}
\caption{\textbf{Lab: } Surface coverage reached in the lab, depending on the different staring positions (A,B,C,D) and the number of robot base positions used. The coverage after the initial explorative scan is also given}

\label{tab:lab_NumPosVsCoverage}%
\begin{tabular}{lllll}
    \toprule
    \textbf{} & \multicolumn{4}{c}{ \textbf{Number of base positions} }\\
    \toprule
    \textbf{Starting Area} & \textbf{Explorative Scan} & \textbf{1} & \textbf{2} & \textbf{3} \\
    \midrule
    A  &$53.42 \pm 15.78$ &$75.79 \pm 8.31$ &$90.59 \pm 6.81$ &$94.69 \pm 6.02$\\
    B  &$54.96 \pm 10.57$ & $70.22 \pm 11.71$ &$89.77 \pm 8.35$ &$97.29 \pm 0.65$\\
    C &$58.63 \pm 4.39$ &$76.83 \pm 3.32$ &$89.04 \pm 6.99$ &$98.11 \pm 0.50$\\
    D &$53.62 \pm 5.61$ &$70.98 \pm 11.45$ &$92.04 \pm 4.27$ &$96.97 \pm 0.95$\\
    
    \midrule
    mean & $57.20 \pm 6.73$& $73.74 \pm 10.12$ &$90.96 \pm 6.45$ &$96.90 \pm 3.16$ \\
    \botrule
    \end{tabular}
\end{minipage}
\end{table}

\section{Discussion}
Our study demonstrates the capability of our mobile robot to perform autonomous surface scanning independent of body geometry and environmental settings.

\subsection{Configuration Space and Path Evaluation}

Our results demonstrate that using three robot base positions is generally sufficient to achieve good coverage across various experimental setups.
For all body models, we achieve comparable results with a coverage of  \SI{93.80 \pm 3.15}{\%}, indicating the systems robustness to variations in body geometry.
Note, that a coverage of \SI{100}{\%} would not be feasible because body parts in confined spaces, such as between the arms and legs and the torso, are impossible to access.
Our results further indicate, that assuming a half-cylinder body model for path planning instead of detailed body geometries is reasonable for all body geometries studied. 
Therefore, we only need to perform a full configuration space analysis once with a half-cylinder body geometry approach, which enhances the real-time capability of our system.

No significant influence of the couch height on the scanning coverage is observed. However, it should be considered in the path planning process to avoid collisions between the robot and the corpse, as well as to ensure the use of the optimal depth range of the RGB-D camera. 

The analysis of the workspace constraints shows that a narrow room still allows a scanning coverage of \SI{90}{\%}. However, if access is limited to one side of the couch, body geometries on the opposite side are particularly difficult to scan, resulting in a reduction in coverage to \SI{84.92}{\%}.

\subsection{Real World Evaluation}
While our configuration space analysis indicates an initial coverage of \SI{69.48 \pm 6.49}{\%} for a single robot base position, the initial explorative scan from our lab experiments has a notably lower coverage of \SI{57.20 \pm 6.73}{\%}.
However, note, that the initial scan position is not optimized and a subsequent second scan from an optimized position yields a coverage of \SI{73.74 \pm 10.12}{\%} that is even better than the expected coverage. 
This indicates that optimized positions are helpful to obtain good coverage.
The need and importance of considering the environmental analysis from the initial scan into the subsequent path planning is demonstrated.

Using the optimized scanning path in the lab experiments, we achieved an overall coverage of \SI{96.90\pm 3.16}{\%} across all starting positions. This consistency demonstrates the robustness and flexibility of our scanning system and the independence of the initial position. 

Considering the qualitative analysis of the stitched point cloud in Fig.~\ref{fig:Lab_acquiredPCD_vsHS}, point clouds obtained from a single robot base position show promising results. However, after stitching, overlapping areas between point clouds from multiple positions become slightly blurred, and distance differences to the reference scan appear mainly at the edges of the stitched sub-point clouds. In contrast, the hands-canned reference scan appears smoother, due to extensive manual post-processing using Artec software, also visible comparing Fig.~\ref{fig:handscannerRAW} and Fig.~\ref{fig:handscannerPP}. We avoided such post-processing for the robotic scans in order to preserve the surface irregularities that are critical to our research. 
Therefore, the use of a more advanced and customized stitching algorithm, specifically focused on stitching partial point clouds \cite{Oguz.2024}, could improve the quality of our stitched surface scan. \\

The surface scans acquired at Legal Medicine achieved a coverage of over \SI{90}{\%} compared to the CT surfaces, which indicates its potential for reliable corpse surface scanning. 
In Figure~\ref{fig:pat1kinectScan_single}, a visible bruise on the arm illustrates the system's ability to capture important legal medicine details.
The surface scans show variability in lighting conditions, which could be addressed by adding external light sources to the system.\\
The extracted CT surface shows artifacts in its geometry, especially at the edges, due to the thresholding method used for extraction. This also explains the larger differences after the registration observed at the edges, where these artifacts are primarily concentrated.

\begin{figure}[tb]
\centering
    \includegraphics[width=0.8\linewidth]{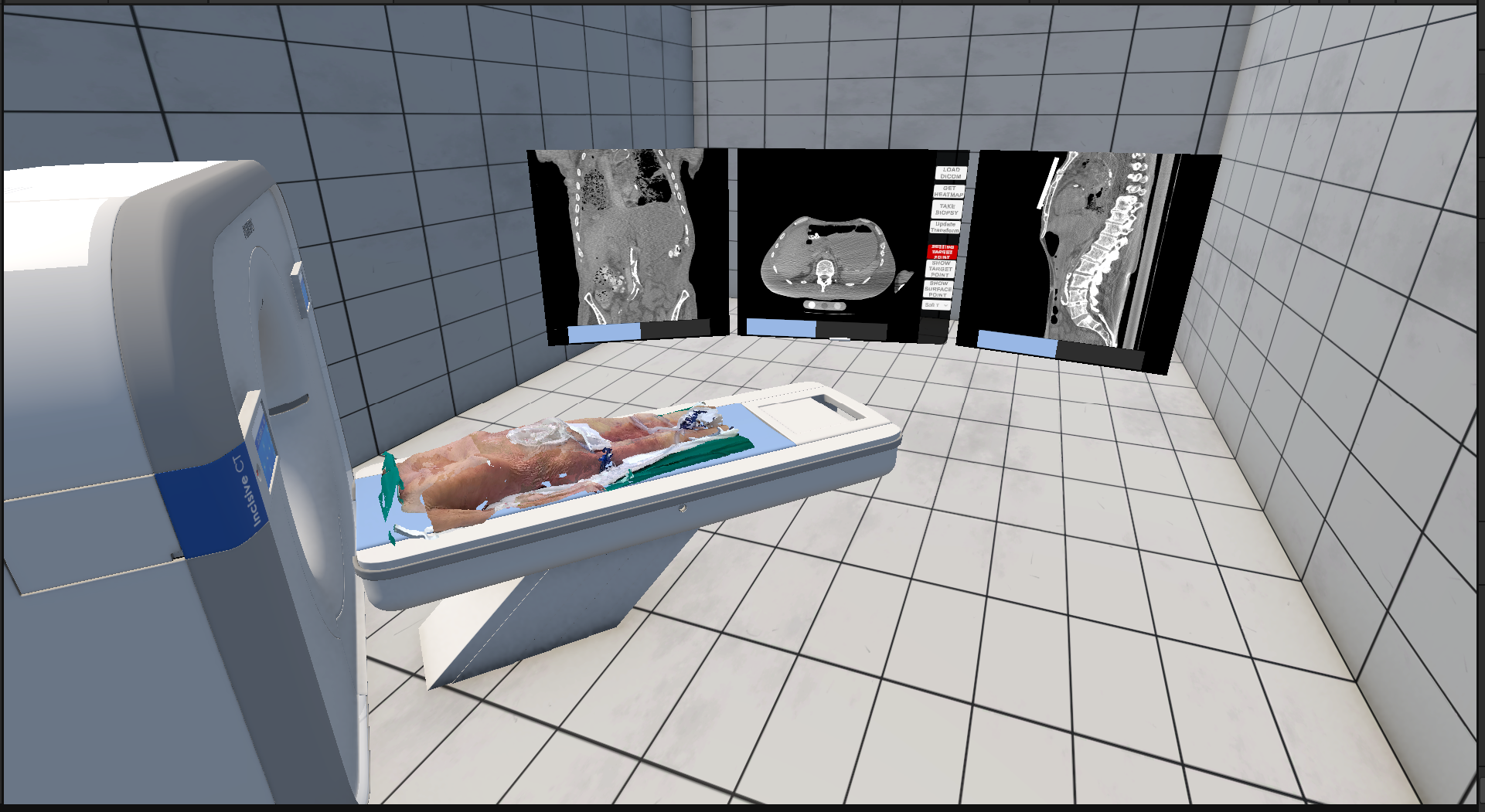}
    \caption{Possible visualization of the scanned surface in the Quest3}
\label{fig:quest3}
\end{figure}

When applying our system in clinical practice, the acquired surface scans could be visualized in a mixed reality platform that combines different imaging modalities, such as the mixed reality platform from brainlab~(Munich, GE).
Figure~\ref{fig:quest3} shows such a visualization in a Quest 3. The acquired surface scan is registered with the corresponding CT scan, allowing visualization of the CT slice corresponding to a particular point on the surface.

Future work on the system could focus on several key enhancements to improve precision and functionality. A more advanced stitching process for multiple point cloud stitching would lead to higher accuracy in spatial models. 
Optimizing the design of robot base is another priority; we aim to create a smaller base with better wheels to improve maneuverability.

The presented robot scanning system shows promising results in autonomously acquiring comprehensive surface scans in different environmental setting using three different robot base positions.

\section{Conclusion}
In this study, we have presented a mobile and autonomous surface scanning robot designed to address the growing need for comprehensive and efficient legal medicine documentation. 
Our system provides an autonomous, cost-effective, and mobile alternative to traditional fixed or manual approaches.
Our results show that the presented robotic scanning system is a promising approach to enable autonomous surface scanning in a variety of environments.  
The conducted analysis underlines the importance of considering environmental settings for calculating the optimal scanning path.
Future work will focus on optimizing the system design to further improve robustness and automation, as well as improving the surface capture device and subsequent stitching process. 
\backmatter




\bmhead{Acknowledgments}
This work was partially funded by the TUHH $i^{3}$ initiative and the Interdisciplinary Competence Center for Interface Research (ICCIR) supported by TUHH and UKE and the European Union under Horizon Europe program (No. 101059903).
\section*{Declarations}
\bmhead{Conflict of interest} S. Grube, S. Latus, M. Fischer, V. Raudonis, A. Heinemann, B. Ondruschka and A. Schlaefer declare that they have no conflict of interest.
\bmhead{Ethical approval} 
This work involved human subjects or animals in its research. Approval of
all ethical and experimental procedures and protocols was granted by Ethics
Committee of the Hamburg Chamber of Physicians under Application No.
2020-10353-BO-ff.
\bmhead{Informed consent} The study is approved by the Ethics Committee of the Hamburg Chamber
of Physicians and the study protocol includes the informed consent of relatives or legal representatives.

\noindent

\bigskip
\bibliography{sn-bibliography}

\end{document}